\documentclass[10pt,twocolumn,letterpaper]{article}

\usepackage{cvpr}

\usepackage{times}
\usepackage{epsfig}
\usepackage{graphicx}
\usepackage{color}
\usepackage{graphics} 
\usepackage{epsfig} 
\usepackage{amsmath} 
\usepackage{amssymb}  
\usepackage{epstopdf}
\usepackage{mathtools}
\usepackage{stfloats}
\usepackage{mathtools}
\usepackage{enumerate}
\usepackage{booktabs}
\usepackage{makecell}
\usepackage{multirow}
\usepackage{amssymb}
\usepackage{pifont}
\newcommand{\cmark}{\ding{51}}%
\newcommand{\xmark}{\ding{55}}%

\newcommand{\PreserveBackslash}[1]{\let\temp=\\#1\let\\=\temp}
\newcommand{\tabincell}[2]{\begin{tabular}{@{}#1@{}}#2\end{tabular}}

\newcolumntype{C}[1]{>{\PreserveBackslash\centering}p{#1}}
\newcolumntype{R}[1]{>{\PreserveBackslash\raggedleft}p{#1}}
\newcolumntype{L}[1]{>{\PreserveBackslash\raggedright}p{#1}}

\newcommand{\HL}[1]{\textcolor[rgb]{0.00,0.00,0.00}{#1}}

\usepackage[pagebackref=true,breaklinks=true,letterpaper=true,colorlinks,bookmarks=false,citecolor=green]{hyperref}

\cvprfinalcopy 

\def\cvprPaperID{3336} 

\ifcvprfinal\fi
\begin{document}

\title{Dense Recurrent Neural Networks for Scene Labeling}

\author{Heng~Fan ~~~ Haibin~Ling\\
Department of Computer and Information Sciences, Temple University, Philadelphia, PA USA\\
{\tt\small \{hengfan,hbling\}@temple.edu}
}

\maketitle
\thispagestyle{empty}

\begin{abstract}
  Recently recurrent neural networks (RNNs) have demonstrated the ability to improve scene labeling through capturing long-range dependencies among image units. In this paper, we propose dense RNNs for scene labeling by exploring various long-range semantic dependencies among image units. In comparison with existing RNN based approaches, our dense RNNs are able to capture richer contextual dependencies for each image unit via dense connections between each pair of image units, which significantly enhances their discriminative power. Besides, to select relevant and meanwhile restrain irrelevant dependencies for each unit from dense connections, we introduce an attention model into dense RNNs. The attention model enables automatically assigning more importance to helpful dependencies while less weight to unconcerned dependencies. Integrating with convolutional neural networks (CNNs), our method achieves state-of-the-art performances on the PASCAL Context, MIT ADE20K and SiftFlow benchmarks.
  \end{abstract}

\section{Introduction}
\label{sec_intro}
\begin{figure}[!htb]
\centering
\begin{tabular}{c}
\includegraphics[width=8cm]{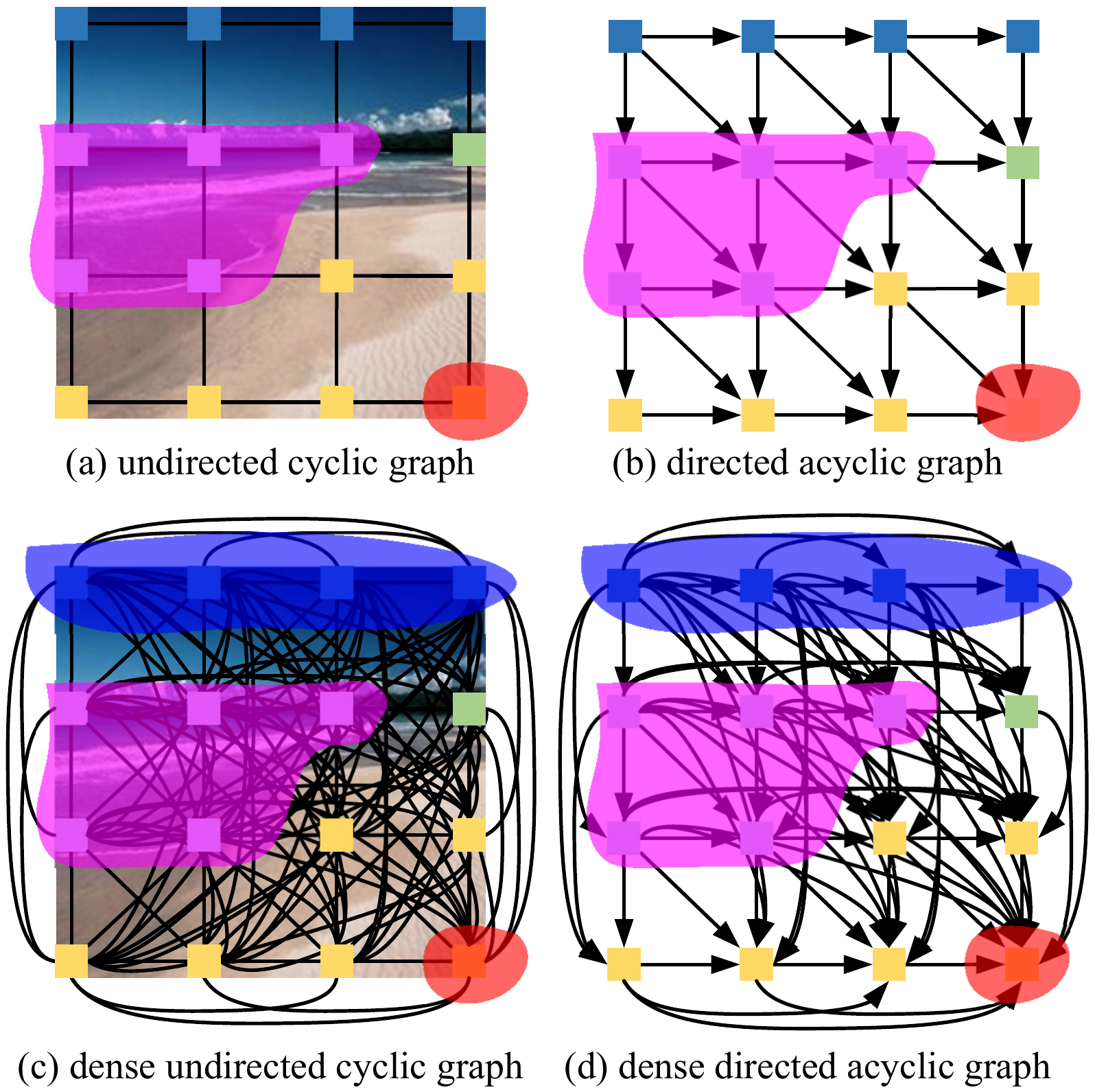}\\
\end{tabular}
\caption{Image (a) shows the image of UCG structure as in~\cite{shuai2017scene}, and image (b) is one of four DAG decompositions. Different from ~\cite{shuai2017scene}, we utilize D-UCG to represent an image as shown in image (c), and image (d) displays one of the four D-DAGs. Compared to UCG and DAG, our D-UCG and D-DAG capture richer dependency information flow in images. Best viewed in color.}
\label{fig:fig1}
\end{figure}
Scene labeling, or scene parsing, which aims to assign one of predefined labels to each pixel in an image, is usually formulated as a pixel-level multi-classification problem. Borrowing from the successes of convolutional neural networks (CNNs)~\cite{lecun1989backpropagation} in image classification~\cite{krizhevsky2012imagenet,simonyan2014very,he2016deep}, there are attempts to apply CNNs on scene labeling~\cite{farabet2013learning,long2015fully,noh2015learning,badrinarayanan2015segnet,girshick2014rich}. Owing to the powerful feature representation of CNNs, these approaches demonstrate promising performance on scene parsing. However, a potential problem with these methods is that CNNs only explore limited contextual cues from a small local field for classification, which is prone to cause misclassifications for visually similar pixels of different
categories. For example, the `sand' pixels can be visually indistinguishable from `road' pixels even for human with limited context. To alleviate this issue, a natural solution is to leverage richer context information to discriminate locally ambiguous pixels~\cite{chen2016deeplab,yu2015multi,liu2015parsenet}. In these approaches, nevertheless, the long-range contextual dependencies among image regions are still not effectively explored, which are crucial in scene parsing.

Motivated by the capacity of capturing long-range dependency among sequential data, recurrent neural networks (RNNs)~\cite{elman1990finding} have recently been employed to model semantic dependencies in images for scene labeling~\cite{byeon2015scene,li2016lstm,shuai2017scene,liang2016semantic,visin2016reseg}, allowing us to perform long-range inferences to discriminate ambiguous pixels.

To model the dependencies among image units, a common way~\cite{shuai2017scene,zuo2016learning} is to represent an image with an undirected cyclic graph (UCG) in which the image units are vertices and their interactions are encoded by undirected edges (see Fig.~\ref{fig:fig1}(a)). Due to the loopy structure of UCGs, however, it is difficult to directly apply RNNs to model dependencies in images. To deal with this problem, an UCG is
approximated with several directed acyclic graphs (DAGs) (see Fig.~\ref{fig:fig1}(b)). Then several DAG structured RNNs are adopted to model the dependencies in these DAGs.

Though these DAG structured RNNs can capture dependencies in images to some extent, quiet a bit of information are discarded. For example in Fig.~\ref{fig:fig1}(a), to correctly distinguish the `sand' unit in red region from the `road' unit, DAG structured RNNs can use the dependency information of `water' units in the pink region from its adjacent neighbors. However, the `water' information may be decaying because it needs to pass through conductors (\ie, the adjacent neighbors of this `sand' unit). Instead, a better way is to directly leverage the dependency information from `water' units to discriminate `sand' unit from `road' unit.

Recently, DenseNet~\cite{huang2016densely} has demonstrated superior performance in image recognition by introducing dense connections to improve information flow in CNNs. Analogous to CNNs, the DAG structured RNNs can be unfolded to a feed-forward network, the dependency information in an image flows from the start vertex at top-left corner to end vertex at bottom-right corner. To incorporate richer dependency information for each image unit, it is natural to add more connections to the RNN feed-forward network as well, as proposed in this paper.

\subsection{Contributions}

Our {\bf first contribution} is to propose dense RNNs, which capture richer dependencies from various abundant connections in images for each image unit. Unlike existing approaches representing an image as an UCG, we formulate each image as a dense UCG (D-UCG), which is a complete graph. In D-UCG, each pair of vertexes are connected with an undirected edge (see Fig.~\ref{fig:fig1}(c)). By decomposing the D-UCG into several dense DAGs (D-DAGs), we propose the DAG structured dense RNNs (DD-RNNs) to model dependencies in images (see Fig.~\ref{fig:fig1}(d)). Compared to plain DAG structured RNNs, our DD-RNNs are able to gain richer dependencies from various levels. For instance in Fig.~\ref{fig:fig1}(c), to correctly recognize the `sand' unit in red region, in addition to the dependencies from its neighbors, our DD-RNNs enable the firsthand use of dependencies from `water' units in the pink region to improve the discriminative power.

The DD-RNNs are able to capture vast dependencies for each image unit through dense connections. For a specific unit, however, certain dependencies are irrelevant to help improve discriminative power. For example in Fig.~\ref{fig:fig1}(d), the `sky' units in blue region are actually not useful to distinguish a `sand' unit in the red region from a `road' unit. Instead, the dependencies from `water' units in the pink region are the most crucial cues to infer its label. Thus, more importance should be assigned to dependencies from `water' units. To this end, we make the {\bf second contribution} by introducing an attention model into DD-RNNs. The attention model is able to automatically select relevant and meanwhile restrain irrelevant dependency information for each image unit, which further enhances their discriminative power.

Last but not least, our {\bf third contribution} is to implement an end-to-end scene labeling system by integrating DD-RNNs with CNNs. For validation, we test the proposed method on three popular benchmarks: PASCAL Context~\cite{mottaghi2014role}, MIT ADE20K~\cite{zhou2017scene} and SiftFlow~\cite{liu2011sift}. In these experiments the proposed method significantly improves the baseline and outperforms other state-of-the-art algorithms. 
The code will be released upon the publication.

The rest of this paper is organized as follows. Section \ref{sec_rel} briefly reviews the related works of this paper. Section \ref{drnns} describes the proposed approach in details. Experimental results are demonstrated in Section \ref{sec_res}, followed by conclusion in Section \ref{sec_con}.

\section{Related Work}
\label{sec_rel}

\subsection{Scene parsing}

As one of the most challenging problems in computer vision, scene parsing has drawn increasing attentions in recent decades. Early efforts mainly focus on the probabilistic graphical model with hand-crafted features~\cite{liu2011sift,tighe2013finding,yang2014context,gould2009decomposing}. Despite great progress, these approaches are restricted due to the use of hand-crafted features.

Inspired by their successes in image recognition~\cite{krizhevsky2012imagenet,simonyan2014very,he2016deep}, deep CNNs have been extensively explored for scene parsing. Long {\it et al.}~\cite{long2015fully} propose an end-to-end scene labeling method by transforming standard CNNs for classification into fully convolutional networks (FCN), resulting in significant improvement from conventional methods. To generate desired full-resolution predictions, various approaches are proposed to learn to upsample low resolution feature maps to high-resolution feature maps for final prediction~\cite{noh2015learning,badrinarayanan2015segnet,lin2016refinenet}. In order to alleviate boundary problem of predictions, graphical models such as Conditional Random Field (CRF) or Markov Random Field (MRF) are introduced into CNNs~\cite{chen2016deeplab,zheng2015conditional,liu2015semantic}. As a pixel-level classification problem, context plays a crucial role in scene labeling to distinguish visually similar pixels of different categories. The work of~\cite{yu2015multi} proposes to introduce the dilated convolution into CNNs to gather multi-scale context for scene labeling. Liu {\it et al.}~\cite{liu2015parsenet} suggest an additional branch in CNNs to incorporate global context for scene parsing.

\subsection{RNNs on computer vision}

Recently, owing to the ability to model spatial dependencies among different image regions, RNNs~\cite{elman1990finding} have been applied to many computer vision tasks such as image completion~\cite{oord2016pixel}, handwriting recognition~\cite{graves2009offline}, image classification~\cite{zuo2016learning} and so forth. Taking into consideration the importance of spatial contextual dependencies in images to distinguish ambiguous pixels, there are attempts to applying RNNs for scene labeling.

The work of~\cite{byeon2015scene} explores the two-dimensional long short term memory (LSTM) networks for scene parsing by taking into account the complex spatial dependencies of pixels in an image. In~\cite{stollenga2015parallel}, Stollenga {\it et al.} introduce a parallel multi-dimensional LSTM for image segmentation. Liang {\it et al.}~\cite{liang2016semantic} propose a graph based LSTM to model the dependencies among superpixels in images. Visin {\it et al.}~\cite{visin2016reseg} suggest to utilize multiple linearly structured RNNs to model horizontal and vertical dependencies in images for scene labeling. Li {\it et al.}~\cite{li2016lstm} extend this method by replacing RNNs with LSTM and apply it to RGB-D scene labeling. Qi~\cite{qi2016hierarchically} proposes to use gated recurrent units (GRUs) to model long-range context. Especially, to exploit more spatial dependencies in images, Shuai {\it et al.}~\cite{shuai2017scene} propose to represent an image with an UCG. By decomposing UCG into several DAGs, they then propose to use DAG structured RNNs to model dependencies among image units.

Different from the aforementioned approaches, we propose dense RNNs to model richer long-range dependencies in images from dense connections, which significantly improves the discriminative power for each image unit.

\subsection{Attention model}

The attention-based model, being successfully applied in Natural Language Processing (NLP) such as machine translation~\cite{bahdanau2014neural}, sentence summarization~\cite{rush2015neural} and so on, has drawn increasing interest in computer vision. Xu {\it et al.}~\cite{xu2015show} propose to leverage an attention model to find out regions of interest in images which are relevant in generating next word. In~\cite{chen2016attention}, Chen {\it et al.} propose scale attention model for semantic segmentation by adaptively merging outputs from different scales. In~\cite{abdulnabi2017episodic}, the attention model is utilized to assign importance to different regions for context modeling in images. The work of~\cite{lu2016hierarchical} introduces co-attention model for question answering. Chu {\it et al.}~\cite{chu2017multi} propose to utilize attention model to combine multi-context for human pose estimation.

To the best of our knowledge, our work is the first to leverage the attention model in RNNs for scene labeling. Our attention model automatically selects relevant and restrains irrelevant dependencies for image units from dense connections, further improving their discriminability.

\section{The Proposed Approach}
\label{drnns}

In this section, we describe the proposed approach in details. Section \ref{sec30} briefly reviews the DAG structured RNNs. Section \ref{sec31} introduces dense RNNs to capture richer dependencies in images. The attention model is applied to dense RNNs in Section \ref{sec32}. Section \ref{sec33} describes the full labeling system by integrating dense RNNs with CNNs.

\subsection{Review of DAG structured RNNs}
\label{sec30}

The linear RNNs in~\cite{elman1990finding} are developed to handle sequential data tasks. Specifically, a hidden unit $h_t$ in RNNs at time step $t$ is represented with a non-linear function over current input $x_t$ and hidden layer at previous time step $h_{t-1}$, and the output $y_t$ is connected to the hidden unit $h_t$. Given an input sequence $\{x_t\}_{t=1,2,\cdots,T}$, the hidden unit and output at time step $t$ can be computed with
\begin{align}
h_t = &\phi(Ux_t+Wh_{t-1}+b) \label{eq1} \\
y_t = &\sigma(Vh_t+c)
\end{align}
where $U$, $V$ and $W$ represent the transformation matrices, $b$ and $c$ are bias terms, and $\phi(\cdot)$ and $\sigma(\cdot)$ are non-linear functions, respectively. Since the inputs are progressively stored in the hidden layers as in Eq. (\ref{eq1}), RNNs are capable of preserving the memory of entire sequence and thus capture long-range contextual dependencies in sequential data.

For an image, the interactions among image units can be formulated as a graph in which the dependencies are forwarded through edges. The solution in~\cite{shuai2017scene} utilizes a standard UCG to represent an image (see again Fig.~\ref{fig:fig1}(a)). To break the loopy structure of UCG, \cite{shuai2017scene} further proposes to decompose the UCG into four DAGs along different directions (see Fig.~\ref{fig:fig1}(b) for a southeast example).

\begin{figure}[!t]
\centering
\begin{tabular}{c}
\includegraphics[width=8.1cm]{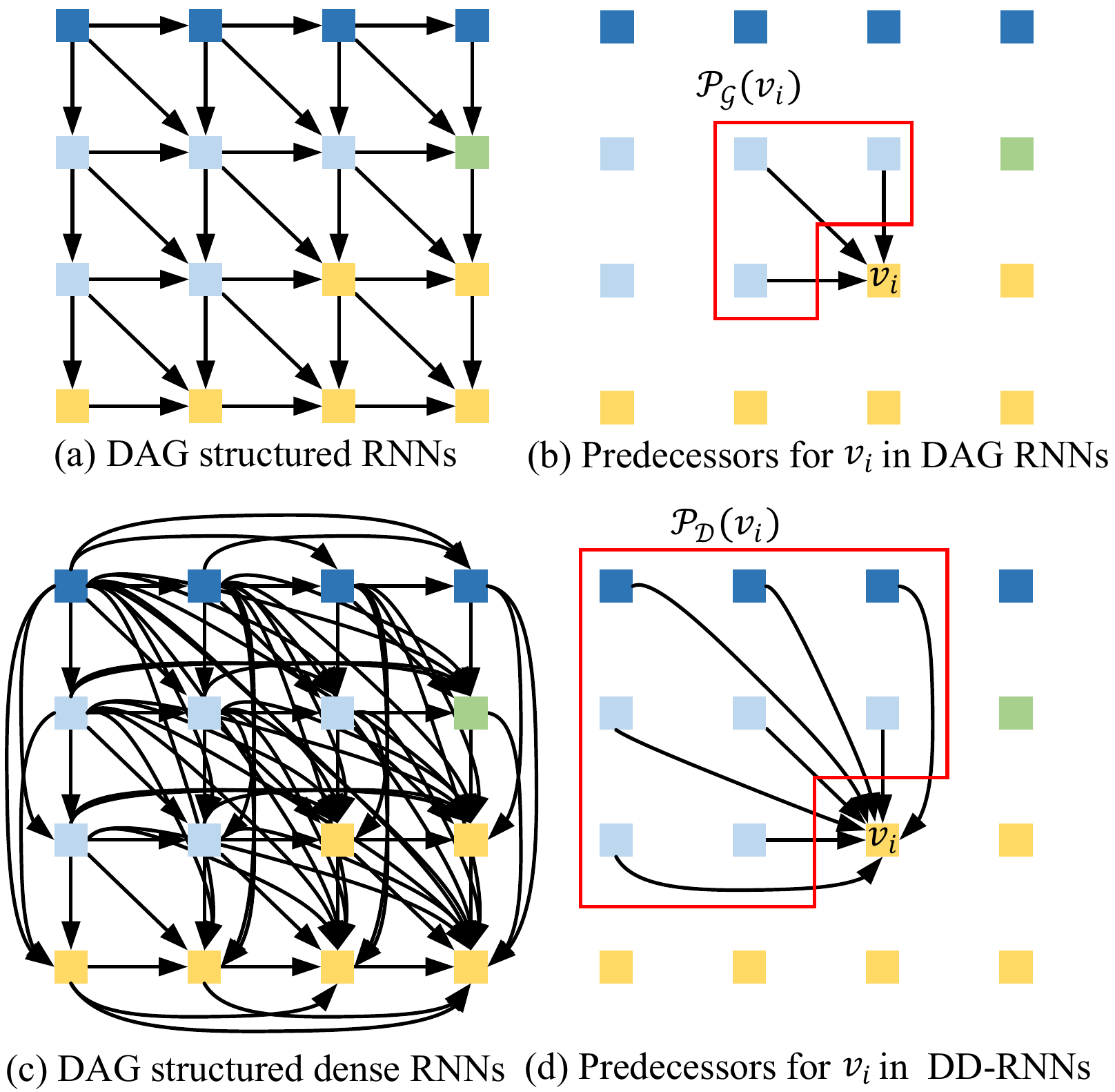}\\
\end{tabular}
\caption{The illustration of difference between DAG structured RNNs~\cite{shuai2017scene} and our DD-RNNs. Image (a) shows the DAG structured RNNs along southeast direction, and in image (b) the hidden layer of vertex $v_i$ relies on its three adjacent predecessors (see the red region in image (b)). Image (c) is our DD-RNNs, and in image (d) the hidden layer of $v_i$ is dependent on all its adjacent and non-adjacent predecessors (see the red region in image (d)). Best viewed in color.}
\label{fig:fig2}
\end{figure}

Let $\mathcal{G}=\{\mathcal{V}, \mathcal{E}\}$ represent the DAG in Fig.~\ref{fig:fig1}(b), where $\mathcal{V}=\{v_{i}\}_{i=1}^N$ denotes the vertex set of $N$ vertices, $\mathcal{E}=\{e_{ij}\}_{i,j=1}^N$ represents the edge set, and $e_{ij}$ indicates a directed edge from $v_i$ to $v_j$. A DAG structured RNN resembles the identical topology of $\mathcal{G}$, with a forward pass formulated as traversing $\mathcal{G}$ from the start vertex. In such modeling, the hidden layer of each vertex is dependent on the hidden units of its adjacent predecessors, (see Fig.~\ref{fig:fig2}(b)). For vertex $v_i$, its hidden layer $h_{v_i}$ and output can be expressed as
\begin{align}
h_{v_i} = &\phi(Ux_{v_i}+W~~\sum\limits_{\mathclap{v_j\in{\mathcal{P}_{\mathcal{G}}(v_i)}}}{h_{v_j}}+b) \label{eq4} \\
y_{v_i} = &\sigma(Vh_{v_i}+c)
\end{align}
where $x_{v_i}$ denotes the local feature at vertex $v_i$ and $\mathcal{P}_{\mathcal{G}}(v_i)$ represents the predecessor set of $v_i$ in $\mathcal{G}$. By storing local inputs into hidden layers and progressive forwarding among them with Eq. (\ref{eq4}), the discriminative power of each image unit is improved with dependencies from other units.

\subsection{Dense RNNs}
\label{sec31}

In DAG structured RNNs, each image unit receives the dependencies from other units through recurrent information forwarding between adjacent units. Nevertheless, the useful dependency information may be potentially degraded after going through many conductors, resulting in a dependency decaying problem. For instance in Fig.~\ref{fig:fig1}(b), the most useful contextual cues from `water' units have to pass through conductors to arrive at the `sand' unit covered in red region. A natural solution to remedy the problem of dependency decaying is to add additional pathes between hidden layers of distant units and current image unit.

Inspired by the recent state-of-the-art DenseNet~\cite{huang2016densely}, we propose DAG structured dense RNNs (DD-RNNs) to model richer dependencies in images. In DenseNet~\cite{huang2016densely}, each layer is connected to every other layer in a feed-forward fashion, which improves information flow between layers. Analogous to CNNs, the DAG structured RNNs can be unfolded to a feed-forward network, the dependency information in an image flows from start vertex at top-left corner to end vertex at bottom-right corner. To capture richer dependencies in images, we introduce more connections in the RNN feed-forward network, resulting in the proposed DD-RNNs.

To achieve dense connections, we in this paper represent an image with a D-UCG, which is equivalent to a complete graph (see Fig.~\ref{fig:fig1}(c) for illustration). Compared to standard UCG, the D-UCG allows each image unit to connect with all of other units. Because of the loopy property of D-UCG, we adopt the strategy as in~\cite{shuai2017scene} to decompose the D-UCG to four D-DAGs along four directions. One of the four D-DAGs along southeast direction is shown in Fig.~\ref{fig:fig1}(d).

Let $\mathcal{D}$ represent the D-DAG in Fig.~\ref{fig:fig1}(d). The structure of our DD-RNNs resembles the identical topology of $\mathcal{D}$ as shown in Fig. \ref{fig:fig2}(c). In our DD-RNNs, the hidden layer of each vertex is dependent on the hidden units of its all \emph{adjacent} and \emph{non-adjacent} predecessors, which fundamentally varies from~\cite{shuai2017scene} in which the hidden unit of each vertex only relies on hidden units of its adjacent predecessors (see Fig.~\ref{fig:fig2}(b)). The forward pass at $v_i$ in DD-RNNs is expressed as
\begin{align}
\hat{h}_{v_i} = &\sum\limits_{\mathclap{v_j\in{\mathcal{P}_{\mathcal{D}}(v_i)}}}{h_{v_j}} \label{eq33}\\
h_{v_i} = &\phi(Ux_{v_i}+W\hat{h}_{v_i}+b) \label{eq44} \\
y_{v_i} = &\sigma(Vh_{v_i}+c) \label{eq55}
\end{align}
where $\mathcal{P}_{\mathcal{D}}(v_i)$ is the dense predecessor set of $v_i$ in D-DAG $\mathcal{D}$, and it contains both adjacent and non-adjacent predecessors (see Fig.~\ref{fig:fig2}(d)). Compared to the DAG structured RNNs in~\cite{shuai2017scene}, our DD-RNNs are able to model richer dependencies in images through various dense connections.

\HL{A concern arisen naturally from the dense model is the complexity. In fact, it is unrealistic to directly apply the DD-RNN to pixels of an image. Fortunately, neither is it necessary. As described in Section~\ref{sec33}, we typically apply DD-RNN to a high layer output of existing CNN models. Such strategy largely reduces the computational burden -- as summarized in Table~\ref{tab:tab5}, our final system runs faster than state-of-the-arts while achieving better labeling accuracies.}

\subsection{Attention model in DD-RNNs}
\label{sec32}

For the hidden layer at vertex $v_i$, it receives dependency information from various predecessors through dense connections. However, the dependency information from different predecessors are not always equally helpful to improve discriminative representation. For example, to distinguish `sand' units from visually alike `road' units in a beach scene image, the most important contextual cues are probably the dependencies from `water' units instead of other units such as `sky' or `tree'. In this case, we term the relation from `water' units as relevant dependencies while the information from `sky' or `tree' units as irrelevant ones.

\begin{figure*}[!htb]
\centering
\begin{tabular}{c}
\includegraphics[width=\linewidth]{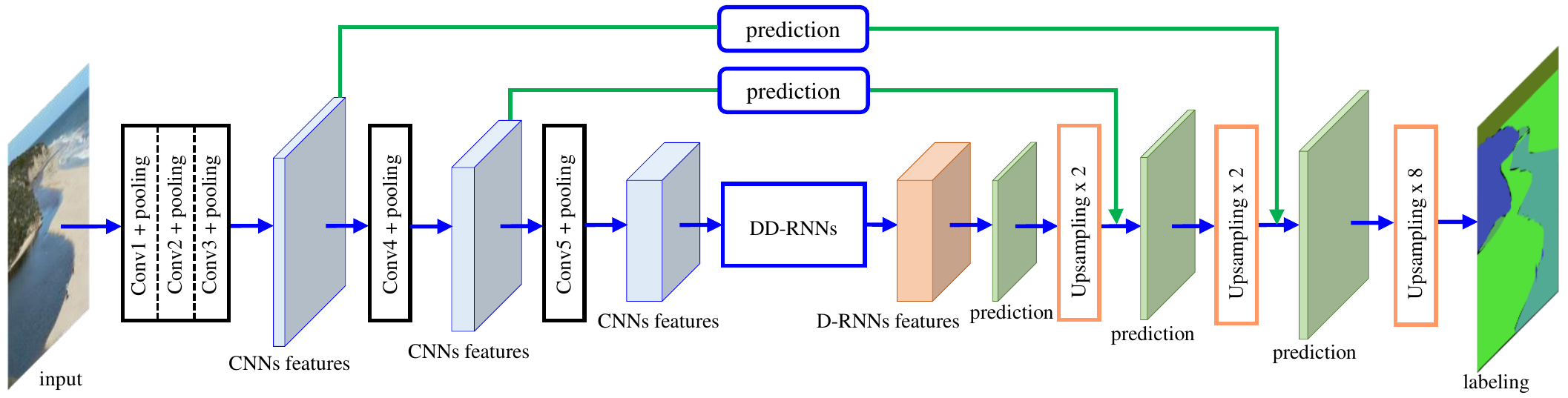}\\
\end{tabular}
\caption{The architecture of the proposed full labeling system. The DD-RNNs are placed on the top of feature maps obtained from the 5$^{\mathrm{th}}$ convolutional block to model long-range dependencies in image, and the deconvolution is used to upsample the predictions. Low-level and high-level features are combined through skip strategy for final labeling (see the green arrows). Best viewed in color.}
\label{fig:fig3}
\end{figure*}

To encourage relevant and restrain irrelevant dependencies for each image unit, we introduce a soft attention model~\cite{bahdanau2014neural} into DD-RNNs. In~\cite{bahdanau2014neural}, the attention model is employed to softly assign importance to input words in a sentence when predicting a target word for machine translation. In this paper, we leverage attention model to select more relevant and useful dependencies for image units. To this end, we do not use Eq. (\ref{eq33}) and Eq. (\ref{eq44}) to directly model the relationships between $h_{v_i}$ and its all predecessors. Instead, we employ the following expression to model the dependency between $h_{v_i}$ and one of its predecessors $h_{v_j}$
\begin{equation}\label{eq6}
h_{v_i,v_j} = \phi(Ux_{v_i}+Wh_{v_j}+b)
\end{equation}
where $h_{v_j}$ represents the hidden layer of one predecessor $v_j\in\mathcal{P}_{\mathcal{D}}(v_i)$ of $v_i$. The $h_{v_i,v_j}$ in Eq. (\ref{eq6}) models dependency information from $h_{v_j}$ for $h_{v_i}$. The final hidden unit $h_{v_i}$ at $v_i$ is obtained by summating all $h_{v_i,v_j}$ with attention, and mathematically computed with
\begin{equation}\label{eq7}
h_{v_i} = \sum\limits_{\mathclap{v_j\in{\mathcal{P}_{\mathcal{D}}(v_i)}}}{h_{v_i,v_j}w_{v_i,v_j}}
\end{equation}
where the attention weight $w_{v_i,v_j}$ for $h_{v_j}$ reflects the relevance of the predecessor $v_j$ to $v_i$, calculated by
\begin{equation}\label{eq8}
    w_{v_i,v_j} = \frac{\mathrm{exp}(z^{\mathrm{T}}h_{v_i,v_j})}{\sum\limits_{\mathclap{v_k\in{\mathcal{P}_{\mathcal{D}}(v_i)}}} ~\exp(z^{\mathrm{T}}h_{v_i,v_k})}
\end{equation}
where $z^{\mathrm{T}}$ represents a transformation \HL{matrix}.


With the above attention model, we replace Equations (\ref{eq33}) and (\ref{eq44}) with Equations (\ref{eq6}) and (\ref{eq7}) for a forward pass at $v_i$ in DD-RNNs. With standard stochastic gradient descent (SGD) method, the attentional DD-RNNs can be trained in an end-to-end manner.

\subsection{Full labeling system}
\label{sec33}

Before showing the full labeling system, we first introduce the decomposition of D-UCG. As in~\cite{shuai2017scene}, we decompose the D-UCG $\mathcal{U}$ into a set of D-DAGs represented with $\{\mathcal{D}^{l}\}_{l=1}^L$, where $L$ is the number of D-DAGs. Since Eq. (\ref{eq7}) only computes the hidden layer at vertex $v_i$ in one of $L$ D-DAGs, the final output $\hat{y}_{v_i}$ at $v_i$ is derived by aggregating the hidden layers at $v_i$ from all D-DAGs. The mathematical formulation for this process is expressed as
\begin{align}
h_{v_i,v_j}^{l} = &\phi(U^{l}x_{v_i}+W^{l}h_{v_j}^{l}+b_l) \label{eq12} \\
h_{v_i}^{l} = &\sum\limits_{\mathclap{v_j\in{\mathcal{P}_{\mathcal{D}^{l}}(v_i)}}}{h_{v_i,v_j}^{l}w_{v_i,v_j}^{l}} \label{eq13}\\
\hat{y}_{v_i} = &\sigma(\sum\nolimits_{l=1}^{L}V^{l}h_{v_i}^{l}+c) \label{eq14}
\end{align}
Through the equations above, we can then utilize the proposed DD-RNNs to model abundant dependencies among image units.

We develop an end-to-end scene labeling system by integrating our approach with popular CNNs for scene parsing as shown in Fig.~\ref{fig:fig3}. The first five convolutional blocks, borrowed from the VGG network~\cite{simonyan2014very}, are used to extract high-level features for local regions. The proposed DD-RNNs are placed on the top of feature maps obtained from the 5$^{\mathrm{th}}$ convolutional block to model long-range dependencies in the input image, and the deconvolution operations are used to upsample the predictions. To produce the desired input size of labeling result, we use the deconvolution~\cite{zeiler2011adaptive} to upsample predictions. Taking into consideration both spatial and semantic information for scene labeling, we adopt the skip strategy~\cite{long2015fully} to combine low-level and high-level features. The whole system is trained end-to-end with the pixel-wise cross-entropy loss. \HL{Finally, we apply conditional random field~\cite{krahenbuhl2011efficient} to further polish the results.}

\section{Experimental Results}
\label{sec_res}

\noindent
{\bf Implementation details.} In our full labeling system, the parameters for the five convolutional blocks are borrowed from VGG network~\cite{simonyan2014very}. DD-RNNs are employed to model dependencies among image units in the 5$^{\mathrm{th}}$ pooling layer. The network takes $512\times{512}$ images as inputs, and outputs the labeling results with same resolution. When evaluating, the labeling results are resized to the size of original inputs. The dimension of input, hidden and output units for D-RNNs is set to 512. The two non-linear activations $\phi$ and $\sigma$ are {\it ReLU} and {\it softmax} functions, respectively. The full networks are end-to-end trained with standard SGD method. For convolutional blocks, the learning rate is initialized to be $10^{-4}$
and decays exponentially with the rate of 0.9 after 10 epochs. For D-RNNs, the learning rate is initialized to be $10^{-2}$ and decays exponentially with the rate of 0.9 after 10 epochs. The batch sizes for both training and testing phases are set to 1. The results are reported after 50 training epoches. The networks are implemented in Matlab using MatConvNet~\cite{vedaldi2015matconvnet} on a single Nvidia GeForce TITAN GPU with 12GB memory.

\noindent
{\bf Evaluation metrics.} In this work, we use three types of metrics, \ie, global pixel accuracy (GPA), average class accuracy (ACA) and mean intersection over union (IoU), to evaluate the proposed method. For details of these metrics, readers are referred to~\cite{long2015fully}.

\noindent
{\bf Baseline.} To better analyze the proposed method, we develop a baseline by using plain DAG structured RNNs to model dependencies. It is worth noticing that the baseline varies from~\cite{shuai2017scene} because we do not use class weighting strategy and larger conventional kernel in our labeling system.

\subsection{Results on PASCAL Context}

The PASCAL Context~\cite{mottaghi2014role} dataset consists of 10,103 images. Following the split in~\cite{mottaghi2014role}, 4,998 images are used for training and the rest for testing. The images are collected from the PASCAL VOC 2010 dataset and re-labeled into 540 classes for pixel-wise scene labeling. Similar to other literatures, we in this paper only consider the most frequent 59 classes in the benchmark for evaluation.

\renewcommand\arraystretch{1.0}
\begin{table}[htbp]\small
  \centering
  \caption{Quantitative results and comparisons on PASCAL Context~\cite{mottaghi2014role} (59 classes). For fair comparisons, we only present algorithms which utilize VGG network~\cite{simonyan2014very} for feature extraction.}
    \begin{tabular}{r|ccc}
    \hline
    Algorithm & GPA (\%) & ACA (\%) & IoU (\%) \\
    \hline
    \hline
    O2P~\cite{carreira2012semantic}     & n/a     & n/a     & 18.1 \\
    CFM~\cite{dai2015convolutional}     & n/a     & n/a     & 34.4 \\
    CAMN~\cite{abdulnabi2017episodic}    & 72.1    & 54.3    & 41.2 \\
    PixelNet~\cite{bansal2017pixelnet} & n/a     & 51.5    & 41.4 \\
    FCN-8s~\cite{long2015fully}  & 69.4    & 50.5    & 38.2 \\
    HO-CRF~\cite{arnab2016higher}  & n/a     & n/a     & 41.3 \\
    BoxSup~\cite{dai2015boxsup}  & n/a     & n/a     & 40.5 \\
    ParseNet~\cite{liu2015parsenet} & n/a     & n/a     & 40.4 \\
    ConvPP-8~\cite{xie2016top} & n/a     & n/a     & 41.0 \\
    CNN-CRF~\cite{lin2016efficient} & 71.5    & 53.9    & 43.3 \\
    CRF-RNN~\cite{zheng2015conditional} & n/a     & n/a     & 39.3 \\
    DeepLab~\cite{chen2016deeplab} & n/a     & n/a     & 37.6 \\
    DeepLab-CRF~\cite{chen2016deeplab} & n/a     & n/a     & 39.6 \\
    DAG-RNN~\cite{shuai2017scene} & 72.7    & 55.3    & 42.6 \\
    DAG-RNN-CRF~\cite{shuai2017scene} & 73.2    & 55.8    & 43.7 \\
    \hline
    \hline
    Baseline & 71.1    & 53.5    & 41.3 \\
    DD-RNNs w/o CRF   &  74.8    & 57.6    &  44.9 \\
    DD-RNNs    & {\bf 75.1}    & {\bf 57.7}    & {\bf 45.3} \\
    \hline
    \end{tabular}%
  \label{tab:tab1}%
\end{table}%

The quantitative results and comparisons with state-of-the-art methods are summarized in Table \ref{tab:tab1}. Benefiting from the power of CNNs, the FCN-8s~\cite{long2015fully} achieves promising result with mean IoU of 38.2\%. To alleviate the boundary issue in FCN-8s, the CRF-RNN~\cite{zheng2015conditional} and DeepLab-CRF~\cite{chen2016deeplab1} propose to utilize probabilistic graphical model such as CRF in CNNs, and obtain better performances with mean IoUs of 39.3\% and 39.6\%, respectively. Other approaches such as CAMN~\cite{abdulnabi2017episodic} and ParseNet~\cite{liu2015parsenet} suggest to improve performance by incorporating global context into CNNs and obtain mean IoUs of 41.2\% and 40.4\%. Despite better performance, these methods still ignore the long-range dependencies in images, which are of importance in inferring ambiguous pixels. The method in~\cite{shuai2017scene} employs RNNs to capture contextual dependencies among image units for scene labeling and demonstrates outstanding performance with mean IoU of 42.6\%. Moreover, they utilize CRF to improve the result to 43.7\%. Different from this method, we propose DD-RNNs to model richer dependencies in images. Our baseline method shows the effectiveness of plain DAG structured RNNs (\ie, no dense connections) for scene labeling with mean IoU of 41.3\%. Without any class weighting strategy and post-operations, our DD-RNNs improves the baseline method from 41.3\% to 44.9\%, which outperforms the method in~\cite{shuai2017scene} by 1.2\%, showing the advantage of DD-RNNs. Further, we apply CRF to polish the result and achieve the mean IoU of 45.3\%.

\begin{figure*}[!htb]
\centering
\begin{tabular}{c}
\includegraphics[width=17cm]{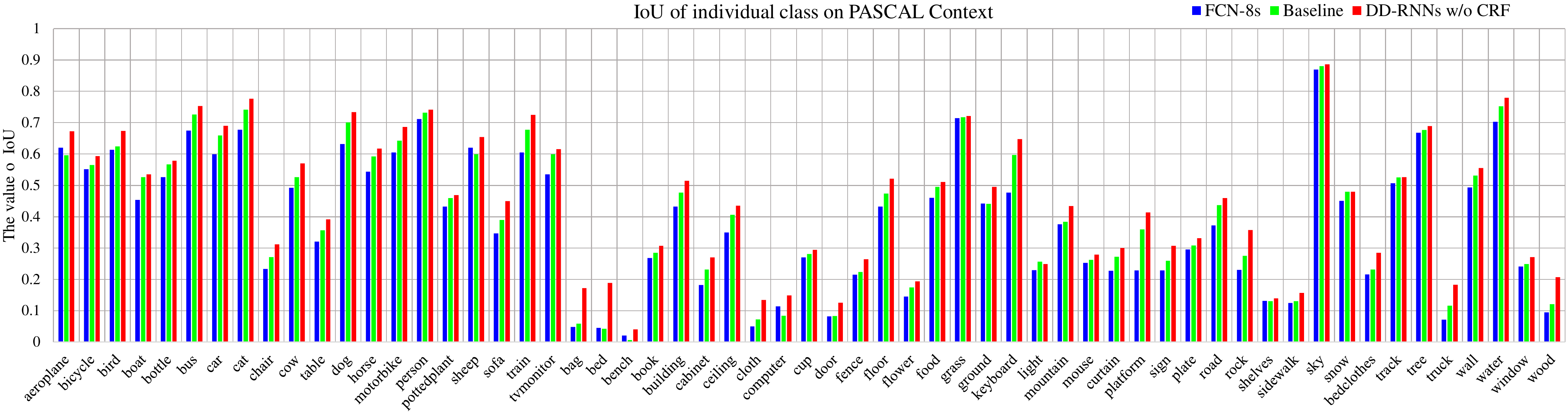}\\
\end{tabular}
\caption{Comparisons of IoU on PASCAL Context~\cite{mottaghi2014role} for each individual class. Best viewed in color.}
\label{fig:fig4}
\end{figure*}

Fig.~\ref{fig:fig4} demonstrates performance comparisons on each individual class in PASCAL Context~\cite{mottaghi2014role} between FCN-8s~\cite{long2015fully} and our methods. From Fig.~\ref{fig:fig4}, we can see that using RNNs in our baseline method can improve the performance most categories including visually similar ones such as `dog' and `horse' with long-range contextual dependencies in images. However, for other similar classes such as `ground' and `sidewalk', the baseline method does not show significant improvements. For few classes such as `bed' and `bench', the FCN-8s~\cite{long2015fully} method even performs better than the baseline. By replacing plain RNNs with dense RNNs, our DD-RNNs achieves significant gains on performance for visually similar classes such as `mountain' and `rock' using richer dependency information.

\begin{figure*}[!htb]
\centering
\begin{tabular}{@{}C{2.83cm}@{}C{2.83cm}@{}C{2.83cm}@{}C{2.83cm}@{}C{2.83cm}@{}C{2.83cm}@{}}
\includegraphics[width=2.75cm]{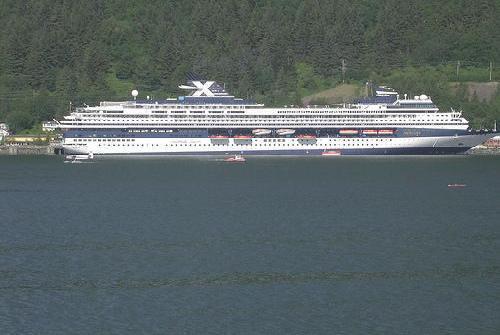}&\includegraphics[width=2.75cm]{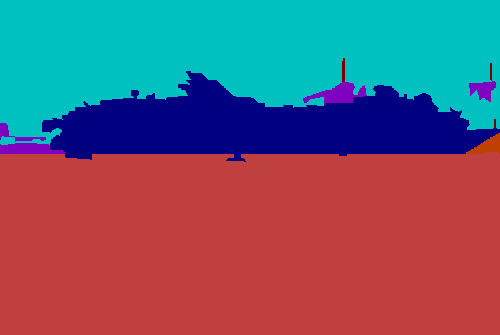}& \includegraphics[width=2.75cm]{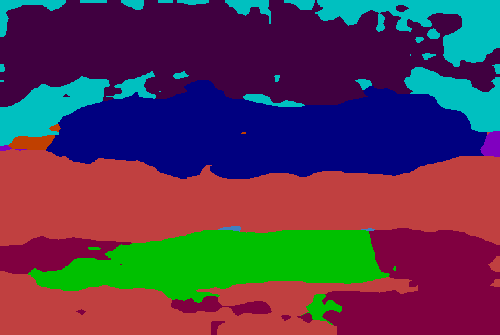}&\includegraphics[width=2.75cm]{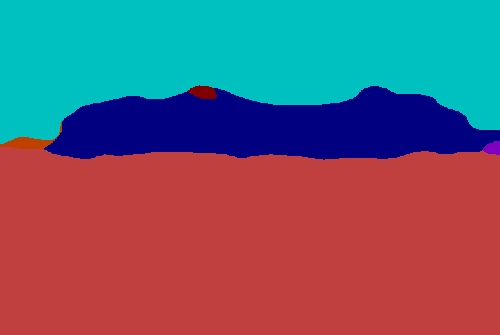}&\includegraphics[width=2.75cm]{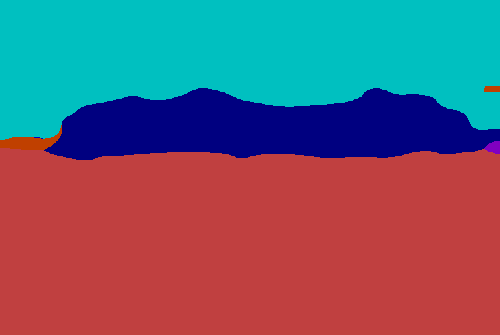}&\includegraphics[width=2.75cm]{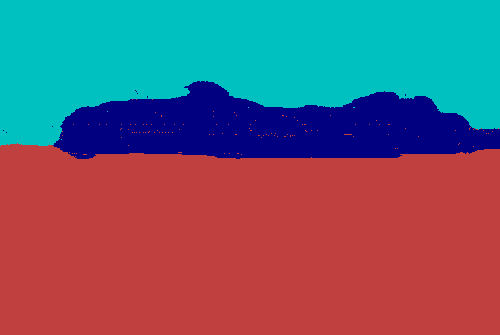}\\
\includegraphics[width=2.75cm]{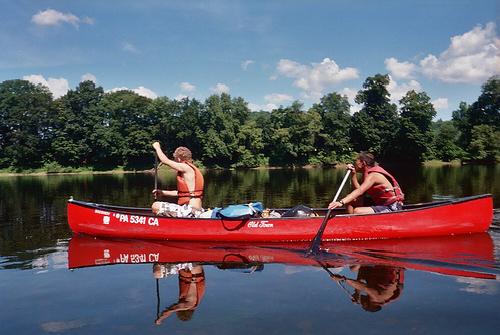}&\includegraphics[width=2.75cm]{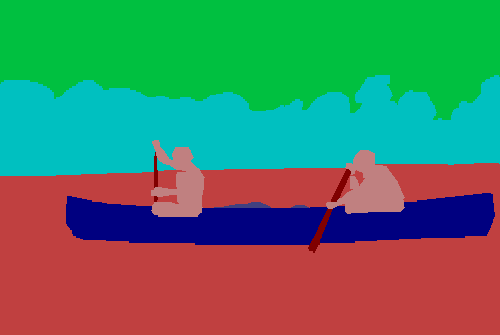}& \includegraphics[width=2.75cm]{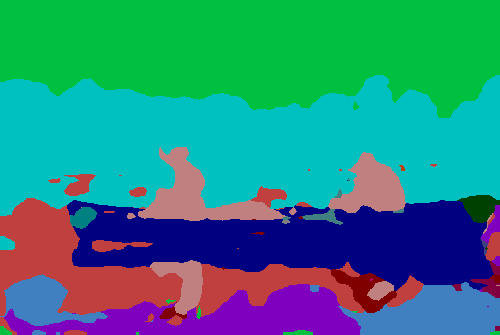}&\includegraphics[width=2.75cm]{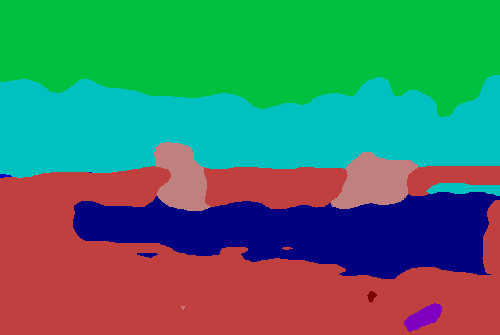}&\includegraphics[width=2.75cm]{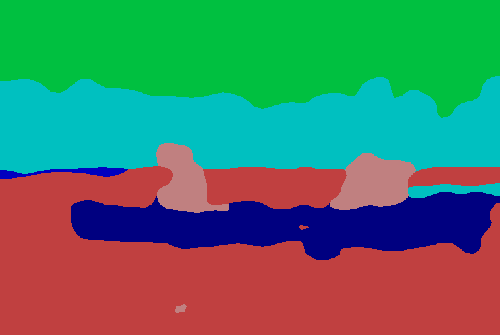}&\includegraphics[width=2.75cm]{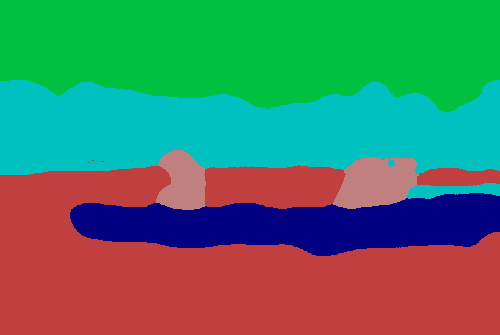}\\
\includegraphics[width=2.75cm]{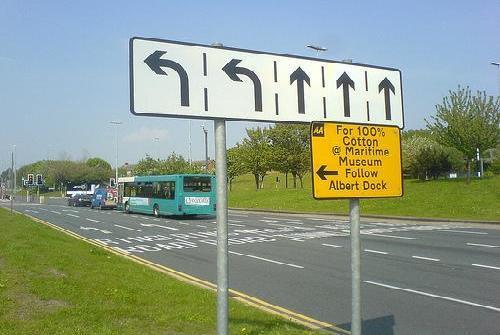}&\includegraphics[width=2.75cm]{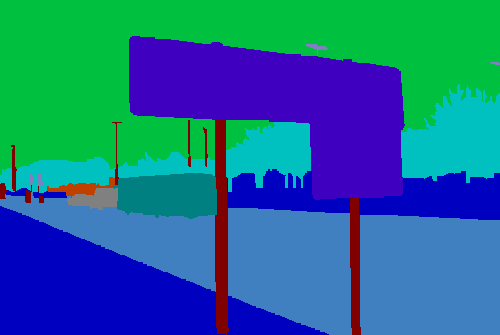}& \includegraphics[width=2.75cm]{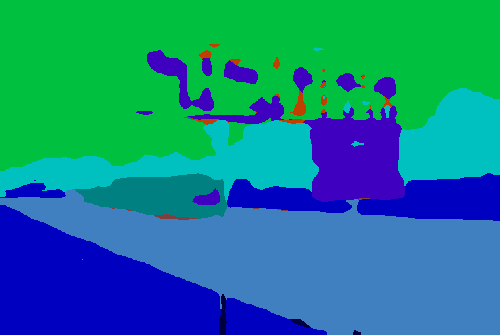}&\includegraphics[width=2.75cm]{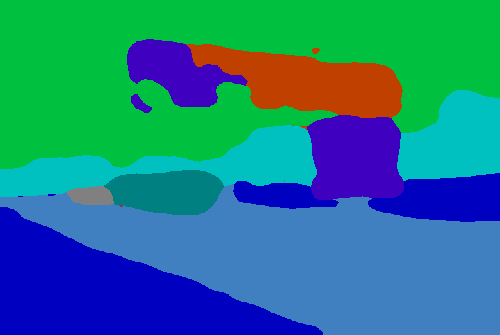}&\includegraphics[width=2.75cm]{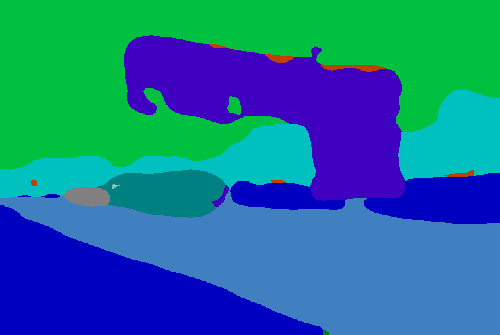}&\includegraphics[width=2.75cm]{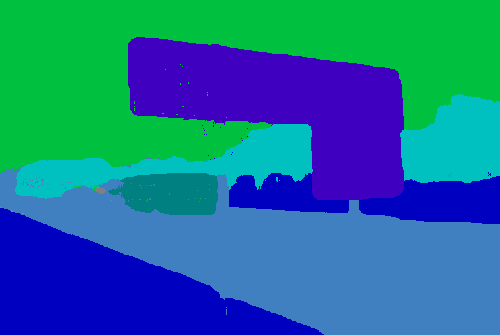}\\
\includegraphics[width=2.75cm]{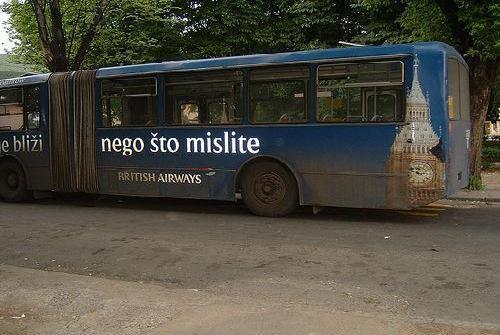}&\includegraphics[width=2.75cm]{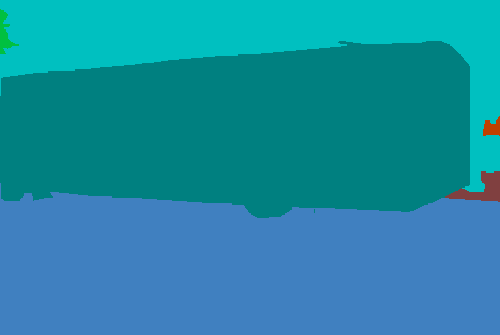}& \includegraphics[width=2.75cm]{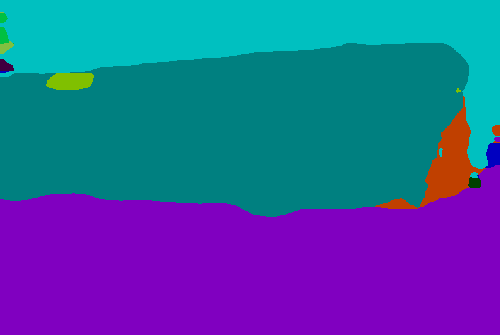}&\includegraphics[width=2.75cm]{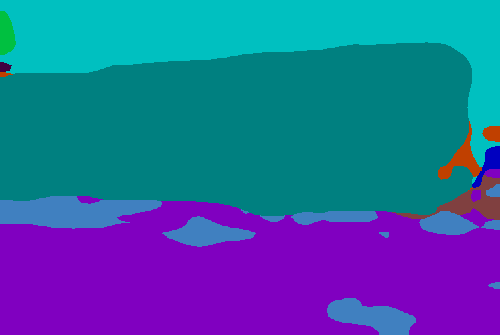}&\includegraphics[width=2.75cm]{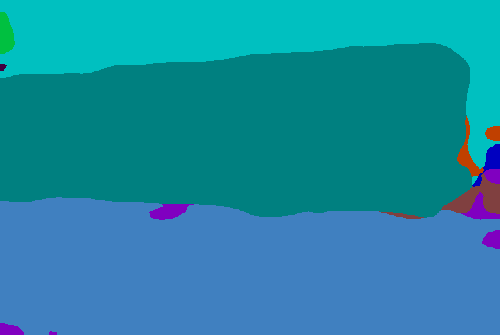}&\includegraphics[width=2.75cm]{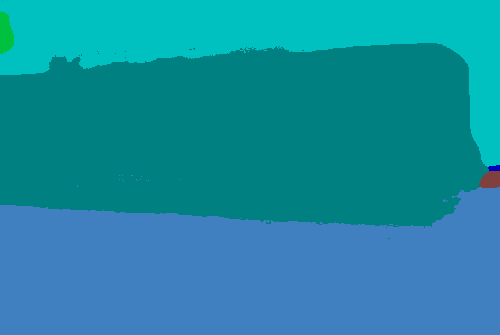}\\
\includegraphics[width=2.75cm]{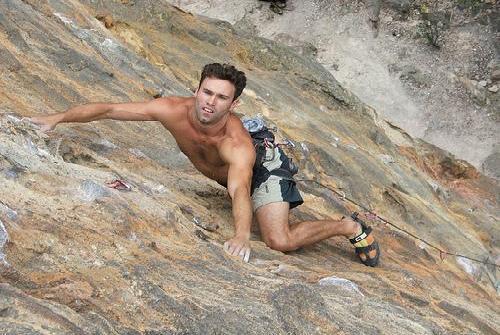}&\includegraphics[width=2.75cm]{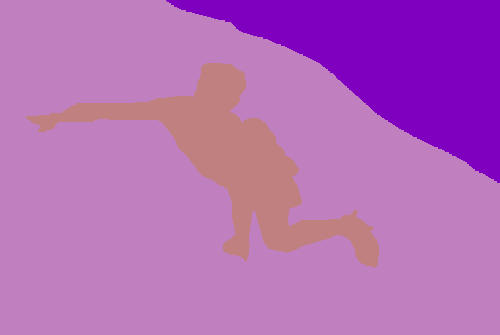}& \includegraphics[width=2.75cm]{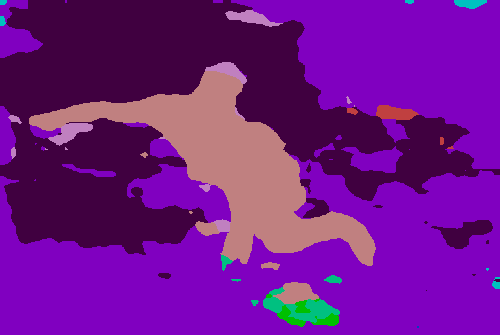}&\includegraphics[width=2.75cm]{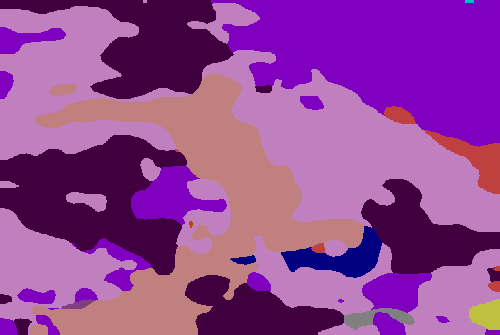}&\includegraphics[width=2.75cm]{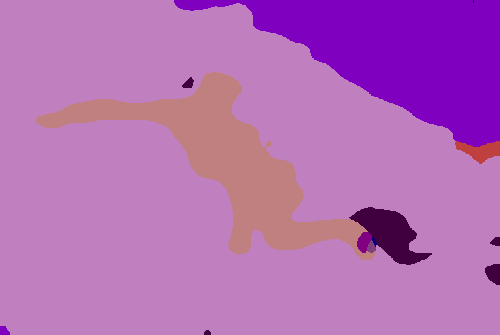}&\includegraphics[width=2.75cm]{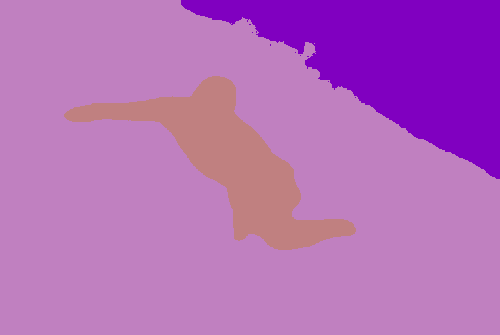}\\
\small{Input} & \small{Groundtruth} & \small{FCN-8s} & \small{Baseline} & \small{DD-RNNs w/o CRF}  & \small{DD-RNNs}\\
\end{tabular}
\caption{Qualitative labeling results on PASCAL Context~\cite{mottaghi2014role}. Best viewed in color.}
\label{fig:fig5}
\end{figure*}

Fig.~\ref{fig:fig5} displays qualitative labeling results on PASCAL Context~\cite{mottaghi2014role}. Without considering long-range contextual dependencies in images, FCN-8s~\cite{long2015fully} is prone to cause misclassifications (see the third column in Fig.~\ref{fig:fig5}). Our baseline approach are able to help alleviate this situation using RNNs to capture dependencies in images. For example, in the first two rows in Fig.~\ref{fig:fig5}, the `water' can be correctly recognized with the dependencies from `boat'. However, the plain RNNs fail in more complex scenes (see the last three rows in Fig.~\ref{fig:fig5}). For example, in the fourth row in Fig.~\ref{fig:fig5}, most of `road' pixels are mistakenly classified into `ground' pixels because of not full use of dependencies from `bus'. By contrast, the proposed DD-RNNs are capable of recognizing most of `road' pixels by taking advantages of richer dependencies from `bus' in images. By employing CRF, the labeling results are further polished (see the last column in Fig.~\ref{fig:fig5}).

\subsection{Results on MIT ADE20K}

The recently published MIT ADE20K~\cite{zhou2017scene} benchmark consists of 20,000 images in training set and 2,000 images in validation set. There are total 150 semantics classes in the dataset. MIT ADE20K~\cite{zhou2017scene} is considered to be one of the most challenging scene parsing benchmarks because of its scene varieties and numerous annotated object instances.

Table \ref{tab:tab2} summarizes the quantitative results and comparisons with other state-of-the-art algorithms. The FCN-8s~\cite{long2015fully} method achieves the result with mean IoU of 29.4\%. To incorporate multi-scale context into CNNs, the work of~\cite{yu2015multi} proposes the dilated convolution and improves the performance to mean IoU of 32.3\%. To the same end, Hung {\it et al.}~\cite{hung2017scene} suggest to embed global context into CNNs to obtain improvements. Based on FCN-8s~\cite{long2015fully} method, they improve the performance to 32.9\% with global spatial prior and to 32.5\% with global feature. Though the aforementioned methods take global context of image into consideration, they still ignore the important contextual dependencies in images. In this work, we employ DD-RNNs to model this dependency information for scene labeling. In our baseline experiment, the plain DAG structured RNNs obtain the result with mean IoU of 32.1\%, which outperforms FCN-8s~\cite{long2015fully} with mean IoU of 29.4\% and SegNet~\cite{badrinarayanan2015segnet} with mean IoU of 21.6\%. By using dense connections in RNNs, the result is further significantly improved to mean IoU of 35.7\% without CRF, demonstrating the effectiveness of dense RNNs.

\renewcommand\arraystretch{1}
\begin{table}[htbp]\small
  \centering
  \caption{Quantitative results on MIT ADE20K~\cite{zhou2017scene}.}
    \begin{tabular}{r|ccc}
    \hline
    Algorithm & GPA (\%) & ACA (\%) & IoU (\%) \\
    \hline
    \hline
    SegNet~\cite{badrinarayanan2015segnet}  & 71.0    & 31.1    & 21.6 \\
    FCN-8s~\cite{long2015fully}  & 71.3    & 40.3    & 29.4 \\
    DilatedNet~\cite{yu2015multi} & 73.6    & 44.6    & 32.3 \\
    FCN-8s+Prior~\cite{hung2017scene} & 75.0    & n/a     & 32.9 \\
    FCN-8s+Feature~\cite{hung2017scene} & 74.5    & n/a     & 32.5 \\
    Cascade-SegNet~\cite{zhou2017scene} & 71.8    & 37.9    & 27.5 \\
    Cascade-Dilated~\cite{zhou2017scene} & 74.5    & 45.4    & 34.9 \\
    \hline
    \hline
    Baseline & 72.8    & 42.6    & 32.1 \\
    DD-RNNs w/o CRF   & 75.6    &  46.9    &  35.7 \\
    DD-RNNs    & {\bf 75.9}    & {\bf 47.3}    & {\bf 36.3} \\
    \hline
    \end{tabular}%
  \label{tab:tab2}%
\end{table}%

\subsection{Results on SiftFlow}

The SiftFlow~\cite{liu2011sift} dataset comprises 2,668 images captured from 8 typical scenes and are annotated with 33 classes. Following the split in~\cite{liu2011sift}, 2,488 images are utilized for training and the rest for testing.

\renewcommand\arraystretch{1}
\begin{table}[htbp]\small
  \centering
  \caption{Quantitative results on SiftFlow~\cite{liu2011sift}.}
    \begin{tabular}{r|ccc}
    \hline
    Algorithm & GPA (\%) & ACA (\%) & IoU (\%) \\
    \hline
    \hline
    RCNN~\cite{liang2015convolutional}    & 83.5    & 35.8    & n/a \\
    RCNN~\cite{liang2015convolutional}    & 79.3    & 57.1    & n/a \\
    FCN-8s~\cite{long2015fully}  & 85.9    & 53.9    & 41.2 \\
    Liu {\it et al}~\cite{liu2011sift}     & 76.7    & n/a     & n/a \\
    ParseNet~\cite{liu2015parsenet} & 86.8    & 52.0    & 40.4 \\
    ClassRare~\cite{yang2014context} & 79.8    & 48.7    & n/a \\
    Tighe {\it et al}~\cite{tighe2013finding}   & 75.6    & 41.1    & n/a \\
    CNN-CRF~\cite{lin2016efficient} & 88.1    & 53.4    & 44.9 \\
    DilatedNet~\cite{yu2015multi} & 86.8    & n/a     & 41.5 \\
    ConvPP-8s~\cite{xie2016top}  & n/a     & n/a     & 40.7 \\
    CNN-LSTM~\cite{byeon2015scene} & 70.1    & 22.6    & n/a \\
    Farabet {\it et al}~\cite{farabet2013learning}     & 72.3    & 50.8     & n/a \\
    Sharma {\it et al}~\cite{sharma2014recursive}     & 75.5     & 48.0     & n/a \\
    DAG-RNN~\cite{shuai2017scene} & 87.3    & 60.2    & 44.4 \\
    DAG-RNN-CRF~\cite{shuai2017scene} & 87.8    & 57.8    & 44.8 \\
    \hline
    \hline
    Baseline & 85.8    & 56.9    & 43.1 \\
    DD-RNNs w/o CRF    &  88.7    &  60.7    & 45.9 \\
    DD-RNNs    & {\bf 89.3}    & {\bf 61.1}    & {\bf 46.3} \\
    \hline
    \end{tabular}%
  \label{tab:tab3}%
\end{table}%

Table \ref{tab:tab3} reports the quantitative results on SiftFlow~\cite{liu2011sift}. The FCN-8s~\cite{long2015fully} approach obtains the result with mean IoU of 41.2\%. By incorporating context, the DilatedNet~\cite{yu2015multi} and CNN-CRF~\cite{lin2016efficient} methods improve the results of mean IoU to 41.5\% and 44.9\%. The work of ~\cite{shuai2017scene} applies DAG structured RNNs to model dependencies in images for scene labeling, and obtains the result with mean IoU of 44.4\%. Furthermore CRF is adopted to refine the results and mean IoU is improved to 44.8\%. Without class weighting strategy or CRF, our DD-RNNs achieve the result with mean IoU of 45.9\%, outperforming the approach in~\cite{shuai2017scene}. Moreover, the result is further improved to 46.3\% with CRF.

\subsection{Ablation study on attention model}

In this paper, we propose the DD-RNNs to model richer dependencies in images, which significantly enhances discriminability for each image unit. However, different dependencies are not always equally helpful. For example, to distinguish a `sand' unit in a beach scene image, the most useful contextual cues are `water' units. Other regions such as `sky' should be paid less attention. To activate relevant and restrain irrelevant dependencies, we introduce an attention model into DD-RNNs. In order to demonstrate the effectiveness of attention model, we conduct experiment by removing attention model from DD-RNNs. Table \ref{tab:tab4} summarizes the experimental results on three benchmarks. From Table \ref{tab:tab4}, we can see that the attention model helps to further improve performance.

\renewcommand\arraystretch{1}
\begin{table}[htbp]\small
  \centering
  \caption{Analysis on the impact of attention model without CRF.}
    \begin{tabular}{@{}C{2.2cm}@{}C{1.8cm}@{}C{1.37cm}@{}C{1.37cm}@{}C{1.37cm}@{}}
    \hline
            & \tabincell{c}{attention \\ model} & GPA (\%) & ACA (\%) & IoU (\%) \\
    \hline
    \hline
    \multirow{2}[0]{*}{Pascal Context~\cite{mottaghi2014role}} & \xmark       & 73.7    & 56.9    & 44.3 \\
            &\cmark       & {\bf 74.8}    & {\bf 57.6}    & {\bf 44.9} \\
    \hline
    \hline
    \multirow{2}[0]{*}{MIT ADE20K~\cite{zhou2017scene}} & \xmark       & 74.8    & 45.7    & 34.5 \\
            & \cmark       & {\bf 75.6}    & {\bf 46.9}    & {\bf 35.7} \\
    \hline
    \hline
    \multirow{2}[0]{*}{SiftFlow~\cite{liu2011sift}} & \xmark       & 88.2    & 59.1    & 45.2 \\
            & \cmark       & {\bf 88.7}    & {\bf 60.7}    & {\bf 45.9} \\
    \hline
    \end{tabular}%
  \label{tab:tab4}%
\end{table}%

\subsection{Study on model complexity}

To further analyze our approach, we show the model size and efficiency in Table \ref{tab:tab5}. The proposed method is based on the VGG network~\cite{simonyan2014very} except for the last fully connected layers. In addition, to generate full input size prediction maps, we utilize three deconvolutional layers to upsample feature maps. Table \ref{tab:tab5} reports the model complexities (\ie, model size and efficiency of one forward pass) and comparisons with other state-of-the-art scene labeling algorithms.

\renewcommand\arraystretch{1}
\begin{table}[htbp]\small
  \centering
  \caption{Analysis of model size and efficiency.}
    \begin{tabular}{rccc}
    \hline
            & \multicolumn{1}{l}{Model Size}  & \multicolumn{1}{l}{Inference} \\
    \hline
    \hline
    SegNet~\cite{badrinarayanan2015segnet}  &    117 MB               &  60 ms\\
    FCN-8s~\cite{long2015fully}  &    497 MB                  &  320 ms\\
    DilatedNet~\cite{yu2015multi} &      513 MB                 &  360 ms\\
    \hline
    \hline
    Ours                           &     190 MB                 & 280 ms \\
    \hline
    \end{tabular}%
  \label{tab:tab5}%
\end{table}%

\section{Conclusion}
\label{sec_con}

In this paper, we propose dense RNNs for scene labeling. Different from existing methods exploring limited dependencies, our DAG structured dense RNNs (DD-RNNs) exploit abundant contextual dependencies through dense connections in image, which better improves the discriminative power of image unit. In addition, considering that different dependencies are not always equally helpful to recognize each image unit, we propose an attention model, which is capable of assigning more importance to relevant dependencies. Integrating with CNNs, we develop an end-to-end scene labeling system. Extensive experiments on three benchmarks including PASCAL Context, MIT ADE20K and SiftFlow demonstrate that our DD-RNNs significantly improve the baseline and outperform other state-of-the-art algorithms, evidencing the effectiveness of dense RNNs.

{\small
\bibliographystyle{ieee}
\bibliography{egbib}
}

\end{document}